\documentclass[10pt,twocolumn,letterpaper]{article}

\usepackage[pagenumbers]{cvpr} 

\definecolor{cvprblue}{rgb}{0.21,0.49,0.74}
\usepackage[pagebackref,breaklinks,colorlinks,allcolors=cvprblue]{hyperref}

\def\paperID{1169} 
\def\confName{CVPR}
\def\confYear{2025}

\newcommand{\method}{SBV\xspace}

\title{Sampling Bag of Views for Open-Vocabulary Object Detection}

\usepackage{colortbl}
\usepackage{caption}
\usepackage{subcaption}
\usepackage{graphicx}
\usepackage{booktabs}
\usepackage{arydshln}
\usepackage{xcolor}         
\usepackage{algorithm}
\usepackage{algpseudocode}
\usepackage{amsmath}
\usepackage{caption}
\usepackage{lineno}
\newcommand{\Hyperparams}{\item[\textbf{Hyperparameters:}]} 
\usepackage{url}
\hypersetup{
    colorlinks=true,
    urlcolor=magenta
}
\usepackage{pifont}
\newcommand{\cmark}{\ding{51}}%
\author{Hojun Choi\\
KAIST AI\\
Republic of Korea\\
{\tt\small hchoi256@kaist.ac.kr}
\and
Junsuk Choe\\
Sogang University\\
Republic of Korea\\
{\tt\small jschoe@sogang.ac.kr}
\and
Hyunjung Shim\\
KAIST AI\\
Republic of Korea\\
{\tt\small kateshim@kaist.ac.kr}
}

\begin{document}
\maketitle

\begin{abstract}
Existing open-vocabulary object detection (OVD) develops methods for testing unseen categories by aligning object region embeddings with corresponding VLM features. A recent study leverages the idea that VLMs implicitly learn compositional structures of semantic concepts within the image. Instead of using an individual region embedding, it utilizes a bag of region embeddings as a new representation to incorporate compositional structures into the OVD task. However, this approach often fails to capture the contextual concepts of each region, leading to noisy compositional structures. This results in only marginal performance improvements and reduced efficiency. To address this, we propose a novel concept-based alignment method that samples a more powerful and efficient compositional structure. Our approach groups contextually related ``concepts'' into a bag and adjusts the scale of concepts within the bag for more effective embedding alignment. Combined with Faster R-CNN, our method achieves improvements of 2.6 box AP$_{50}$ and 0.5 mask AP over prior work on novel categories in the open-vocabulary COCO and LVIS benchmarks. Furthermore, our method reduces CLIP computation in FLOPs by 80.3\% compared to previous research, significantly enhancing efficiency. Experimental results demonstrate that the proposed method outperforms previous state-of-the-art models on the OVD datasets.
\end{abstract}
\addtocontents{toc}{\protect\setcounter{tocdepth}{-1}}
\section{Introduction}
\label{sec:intro}

\begin{figure}[t]
  \centering
   \includegraphics[width=\linewidth]{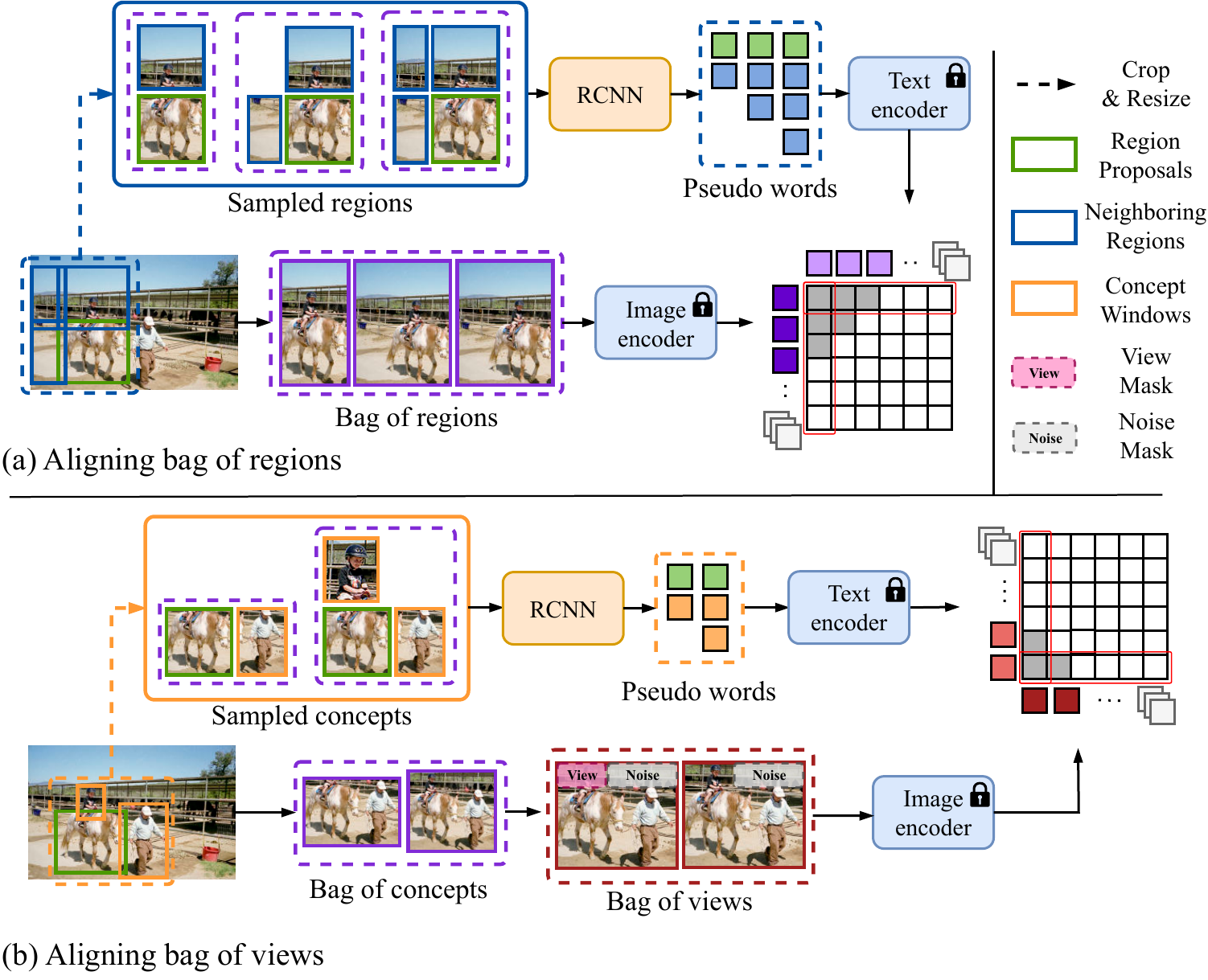}

   \caption{\textbf{(a)} BARON forms a bag of regions by sampling contextually related neighboring regions for each region proposal. It aggregates region embeddings within a bag into pseudo words in the word embedding space and feeds them to the text encoder. This generates a bag-of-regions embedding aligned with the image feature from the frozen VLMs. \textbf{(b)} Our method groups contextually related ``concepts'' into a bag. These concepts are then mapped to views at optimal scales, with masks applied to enhance embedding alignment. Finally, our method aligns the bag-of-view embedding with the corresponding features from the frozen VLMs.}
   \label{fig:1_teaser}
\end{figure}

Open-vocabulary object detection (OVD) methods leverage pre-trained vision-language models (VLMs) \cite{BERT, CLIP, SAM} to recognize classes that were not seen during training. VLMs are trained on large-scale datasets with paired image and text data, making them beneficial for understanding classes that detectors have not encountered during training.

Recently, BARON \cite{BARON} proposed a \textit{bag-of-regions} embedding to better exploit the compositional structures among diverse semantic concepts (\eg contextual co-occurrence between person and airplane) that VLMs are known to learn implicitly \cite{MaskCLIP}. Specifically, BARON: 1) groups contextually related neighboring regions for each region proposal extracted from Faster R-CNN \cite{fasterrcnn} and construct a \textit{bag}; 2) projects regions within the bag into the word embedding space using Faster R-CNN and a linear layer, resulting in pseudo-words; 3) passes pseudo-words through the text encoder to obtain a bag-of-regions embedding; and 4) aligns this bag-of-regions embedding with the VLM’s image encoding during training (\cref{fig:1_teaser}-a). This approach enables BARON to exploit the complex semantic structures inherent in VLMs, achieving significant performance gains.

However, BARON constructs a bag by grouping neighboring regions without considering their semantic meaning. By focusing only on spatial proximity, BARON not only often samples background regions that are simply nearby but also struggles to capture surrounding objects of different sizes and their relationships. This results in reduced efficiency and marginal performance improvements.

In this paper, we propose a novel adaptive sampling strategy that selectively samples surrounding windows by considering semantic information, which we call \textit{concept windows}. The proposed technique captures compositional structures more accurately than BARON, enabling more effective use of the knowledge embedded in VLMs. Specifically, we first obtain an overly large number of candidate concepts. These concepts are represented by proposals with inherent objectness scores from the Region Proposal Network (RPN) \cite{fasterrcnn}. RPN is commonly used in OVD \cite{OVR-CNN, RegionCLIP, BARON} for effectively detecting potential novel objects \cite{vild, detic, bind}. To capture meaningful key concepts, we construct a graph-based canvas where region proposals, filtered through Non-Maximum Suppression (NMS), are nodes and semantic associations form edges. These edges are formed through probabilistic exploration, converging towards dense, high-objectness concepts that are more likely to contain objects. We sample a few useful concepts from these edges to create a \textit{bag of concepts} (\cref{fig:1_teaser}-b).

To consider concepts of varying sizes and their relationships, we sample a \textit{bag of views}, which we refer to as \method (\textbf{S}ampling \textbf{B}ag of \textbf{V}iews). Specifically, we represent concepts in the bag through three distinct perspectives: the global view, which represents the entire image; the local view, which represents each concept individually; and the middle view, which merges multiple concepts. We determine the optimal view based on the size of each view and the number of co-occurring concepts. This enables us to adjust each concept’s scale to better fit the scene. For example, a smaller view is preferred if a broader view excessively increases scene complexity. Conversely, a broader view is preferred if it includes more concepts effectively, as it may better capture the structure. Furthermore, to leverage view embedding alignment, we use separate masks to weigh the importance of each view during CLIP feature extraction.

We conduct extensive experiments on two challenging benchmarks, OV-COCO and OV-LVIS. Our approach achieves 36.6 box AP$_{50}$ on novel categories in OV-COCO and 23.1 mask mAP on novel categories in OV-LVIS, surpassing previous state-of-the-art methods. Notably, our method reduced CLIP computation (FLOPs) by \textbf{80.3}\% compared to BARON, demonstrating its efficiency. 
\section{Related Work}
\label{sec:related}
Open-vocabulary object detection (OVD) aims to detect objects from novel categories unseen during training. Many prior studies leverage large vision-language models (VLMs) \cite{CLIP, SAM}, trained on extensive image-text pairs, to perform zero-shot recognition. The key to improving performance for unseen classes lies in effectively utilizing the alignment between images and text inherent in VLMs.

To achieve this, methods such as \cite{detpro, OV-DETR, CORA} employ prompt modeling to transfer knowledge through learned prompts, enabling more precise contextual descriptions of each class category and thus enhancing performance. Several studies use weakly supervised techniques, such as visual grounding data \cite{GLIP}, image captions \cite{PB-OVD, OVR-CNN, RegionCLIP}, and image labels \cite{vild, detic}. Other approaches \cite{DVDet, SHiNe, ContextDET} reinforce the text modality using Large Language Models (LLMs). In contrast, InstaGen \cite{InstaGen} focuses on the image modality, enhancing novel class prediction by using synthetic images generated by an image generation model. Grounding DINO \cite{groundingdino} is also noteworthy, enabling prompt-based object detection by exchanging information between VLMs and detection transformers \cite{dino} through cross-modality fusion.

While many methods have been proposed, distillation-based methods \cite{vild, OADP, RegionCLIP, SAMP, bind} currently represent one of the highest-performing techniques, aligning individual region embeddings of an object detector with the features extracted from VLMs to leverage VLM knowledge effectively. Recently, BARON \cite{BARON} has advanced beyond using individual regions by grouping neighboring regions around region proposals and aligning them with VLM features. This approach captures the compositional structure of concepts inherent in VLMs and has achieved significant performance gains. 

However, we argue that some of BARON's design choices are suboptimal. Specifically, BARON does not consider semantics when sampling neighboring regions, often resulting in the sampling of unnecessary regions. Consequently, we believe BARON suffers from low computational efficiency and has room for further performance improvement. Our proposed method, in contrast, samples regions with semantic consideration, reducing unnecessary sampling, thereby achieving high computational efficiency and enhanced performance simultaneously.
\section{Preliminaries}
\label{sec:prelim}

\begin{figure}[t]
  \centering
   \includegraphics[width=\linewidth]{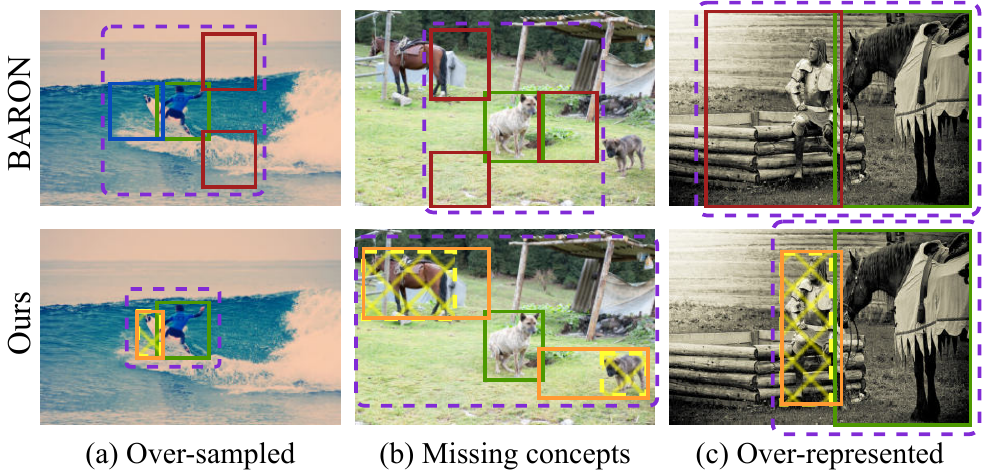}

   \caption{\textbf{Inefficiency of nearest-neighbor sampling in BARON} \cite{BARON}. Red boxes highlight sampling errors and yellow dotted boxes indicate semantic concepts. \textbf{(a)} BARON increases computational costs by over-sampling empty neighbors. \textbf{(b)} Smaller proposals may miss or cut off co-occurring concept boundaries. \textbf{(c)} Larger region crops introduce noise and reduce the relative size of key objects. Our method addresses these issues by sampling concept windows that effectively capture surrounding concepts.}
   \label{fig:2_BARON_limit}
\end{figure}

In this paper, we instantiate the idea of BARON \cite{BARON} to capture compositional structures. BARON samples the nearest neighbors around each region proposal within an image and groups them into a bag. This enables VLMs to infer the compositional structure of semantic concepts. VLMs implicitly learn the compositional structure of complex scenes from large-scale image-text pairs, aligning each concept's pixel embedding with its corresponding text representation. MaskCLIP \cite{MaskCLIP} demonstrates this by showing that VLMs \cite{BERT, CLIP, SAM} effectively capture representations of concepts in complex scenes not explicitly learned. 

\noindent \textbf{Limitations of Bag of Regions.}
However, we argue that BARON’s design has two limitations: 1) BARON constructs a bag by grouping neighboring regions without considering their semantic meaning, reducing its efficiency in sampling important structures within a scene. By relying solely on spatial proximity, BARON often samples nearby backgrounds with no semantic concepts. For example, as shown in \cref{fig:2_BARON_limit}-a, BARON includes backgrounds like the sky, which limits its use of contextual co-occurrence among semantic concepts and increases computational costs.

2) Additionally, since \textit{bags} are created based on fixed window sizes defined by region proposals, BARON struggles to capture surrounding objects of varying sizes and their relationships. As seen in \cref{fig:2_BARON_limit}-b and c, smaller windows may capture only part of a surrounding object, such as a portion of a horse, missing important details, while larger windows may include too irrelevant background around the target object within region proposals.

To quantify this phenomenon, we measure the average number of unnecessary neighboring regions using ground truth (GT) boxes with IoU and noise embeddings with cosine similarity on the COCO benchmark \cite{coco}. For simplicity, we treat GT boxes as semantic concepts and noise embeddings as irrelevant backgrounds. The noise embeddings are derived from BARON's contextual embeddings, which are learnable by training an additional linear layer on the model's class tokens. This setup allows learning features for non-semantic regions, such as backgrounds. As shown in \cref{fig:fig3}, the noise embeddings effectively capture the general characteristics of background areas, such as sidewalks.

In our analysis, neighboring regions extracted by BARON are deemed unnecessary if they overlap less than 85\% IoU with the GT boxes or their embeddings have a cosine similarity score above 0.8 with the noise embeddings. We demonstrate the validity of this setting in \cref{sec:sup:s1}. Consequently, we observe that such unnecessary neighboring regions account for an average of 73\% of all neighboring regions per image. These unnecessary neighbors lead to marginal performance gains and high computational costs when utilizing VLMs. Our method is motivated by the need for a more effective and efficient sampling strategy.
\begin{figure*}[t]
  \centering
   \includegraphics[width=\linewidth]{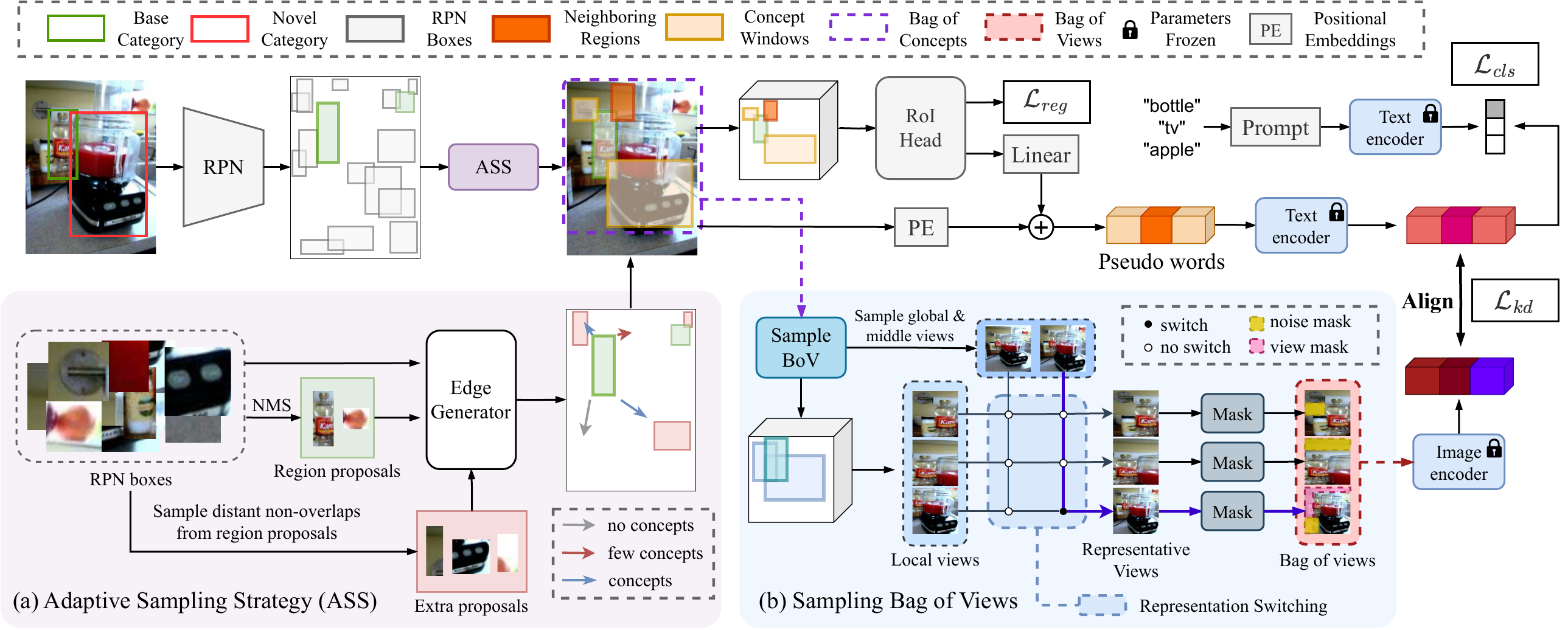}

   \caption{\textbf{Overview of \method}. We call our method \method (\textbf{S}ampling \textbf{B}ag of \textbf{V}iews). \textbf{(a)} \method probabilistically samples a bag from surrounding semantic concepts detected by generating edges on the canvas, where vertices represent region proposals, edges indicate semantically interrelated visual concepts, and coordinates encode the probabilities of nearby RPN boxes. \textbf{(b)} \method selectively obtains a representative view for each concept in the bag from three distinct views (\eg global, middle, and local views) for more effective embedding alignment. These views are further refined using view and noise masks to enhance scene recognition. \textbf{(c)} \method aligns student and teacher embeddings for the bag of views from pre-trained VLMs.}
   \label{fig:main}
\end{figure*}

\section{Methodology}
\label{sec:method}
We call our method \method (\textbf{S}ampling \textbf{B}ag of \textbf{V}iews). Our method aligns embeddings from a \textit{bag of views}, going beyond \textit{individual} regions or a \textit{bag of regions} for OVD (\cref{fig:main}). In this work, we design a powerful and efficient sampling strategy to form a \textit{bag of concepts} that effectively captures semantic concepts around region proposals (\cref{sec:method:m2}). This bag is then used to generate three hierarchical views for each concept in the bag. Using these candidate views, the concepts in the bag are mapped to \textit{views} that best represent optimal scales with masks applied to improve the effectiveness of embedding alignment (\cref{sec:method:m3}).

\subsection{Adaptive Sampling Strategy}
\label{sec:method:m2}
In this section, we introduce an adaptive sampling strategy that creates surrounding windows to capture contextually interrelated semantic concepts around region proposals (\cref{fig:main}-a). Our sampling process is carried out in two steps: 1) identifying surrounding key semantic concepts from RPN boxes (\cref{sec:method:m2:canvas}) and 2) forming a bag of concepts to incorporate these concepts (\cref{sec:method:m2:canvas}). 

\vspace{-2mm}
\subsubsection{Identifying Surrounding Semantic Concepts}
\label{sec:method:m2:canvas}
We focus on extracting semantic concepts around region proposals within an image. We treat RPN boxes, which are widely used in OVD \cite{bind, BARON} and potentially represent novel objects, as candidate concepts. Since the top-$k$ RPN boxes can be noisy and overly abundant (\eg $k$ is 300), we design a canvas to sample a few key semantic concepts around region proposals. This process generates edges on the canvas to explore semantic relationships between the proposals, converging towards denser, high-objectness concepts that are more likely to contain relevant objects (\cref{fig:main}-a).

\noindent \textbf{Canvas construction.}
We first divide each image into a grid with intervals of $\Delta$, referred to as the \textit{canvas}, and generate coordinates for each interval. The canvas encodes the probabilities of RPN boxes in the cardinal directions (\eg up, down, left, and right) around each coordinate, reflecting the likelihood of relevant concepts. To compute this likelihood, we leverage two key properties of RPN \cite{fasterrcnn}: 1) RPN boxes have objectness scores, and 2) RPN boxes are densely populated in areas with objects. To examine the extent of this population, we check the overlap between RPN boxes and a directional $\Delta$-sized box in each cardinal direction at each coordinate. We then calculate the average probability of the top-$k$ RPN boxes by multiplying their objectness score with their IoU against the corresponding directional box in each direction. This yields the probability of relevant concepts for each direction at each coordinate. To speed up the canvas construction, we reduce the top-$k$ RPN boxes from Faster R-CNN \cite{fasterrcnn} to a smaller set of non-overlapping boxes $\boldsymbol{\mathcal{R}}_{\text{reduced}}$, as detailed in \cref{sec:sup:s1:overlap}.

\noindent \textbf{Edge generation.}
Our method explores contextual relationships by traversing the canvas across coordinates, examining all pairs of region proposals obtained from Faster R-CNN. This exploration is guided by coordinates that determine the direction with the highest probability of containing relevant concepts. We model this exploration as edges, which represent the search for key semantic concepts.

\vspace{-0.1mm}
Specifically, an edge is processed by navigating coordinates between a pair of region proposals under the following constraints: 1) edges avoid visiting other region proposals to prevent overfitting to easily identifiable objects within those proposals, 2) probabilities for directions that do not lead to the destination are set to zero, and 3) traversal stops when no valid path remains or when it moves outside the image. To leverage probabilistic variability, we generate $E$ different edges for each pair of proposals. This approach helps improve robustness in sampling semantic concepts. Ultimately, the edges converge to a small set of key visual concepts $\boldsymbol{\nu}$ that are common across all edges for every pair of region proposals in each image. We provide visualizations of our method in \cref{sec:sup:s2:main}, demonstrating its ability to effectively sample key concepts through edges.

However, this design depends on the properties (\eg count and location) of region proposals, meaning that fewer proposals result in fewer edges. To improve the detection of semantic concepts, we greedily select $N$ extra proposals from $\boldsymbol{\mathcal{R}}_{\text{reduced}}$, each chosen to be the farthest, on average, from existing region proposals and previously selected ones. This strategy maximizes edge diversity by encouraging exploration from different directions. More details on this greedy sampling are provided in \cref{sec:sup:s1:fartest}.

\subsubsection{Forming Bag of Concepts}
\label{sec:method:m2:dr}
With the knowledge of visual concepts $\boldsymbol{\nu}$, we can capture diverse contextual relationships between semantic concepts. To leverage this contextual information, we aim to sample visual concepts that provide optimal context for each region proposal. These sampled concepts are referred to as representative concepts $\nu^*$, which default to the neighboring regions in BARON for simplicity.

We adjust the sampling probability according to distance and aspect ratio, as they contribute to contextual information. We believe that increasing the distance between region proposals and visual concepts captures more semantic concepts. Meanwhile, the aspect ratio preserves the original shape of objects during resizing for CLIP input. We sample the representative concept $\boldsymbol{\nu}^*$ for each surrounding direction of region proposals in \cref{eq:eq1}:
\begin{equation}
    \nu^*_{i,j} = \mathop{\arg\max}_{\nu \in \boldsymbol{\nu}_{i,j}} \left( \lambda \cdot \| d_{\nu, i,j} \|_{\eta} + \alpha \cdot \| \gamma_{\nu, i,j} \| \right),
    \label{eq:eq1}
\end{equation}%
where $\lambda$ and $\alpha$ are hyperparameters for distance and aspect ratio, $\| \cdot \|_{\eta}$ denotes normalization by $\eta$, $d_{\nu, i,j}$ computes a distance between a visual concept $\nu$ and the $i$-th region proposal $\mathcal{R}_i$, and $\gamma_{\nu, i,j}$ computes the merged box ratio between $\nu$ and the $j$-th neighbor of the $i$-th region proposal $\mathcal{R}_i$.

To enhance cost efficiency, we discard surrounding windows around region proposals that lack at least one visual concept $\boldsymbol{\nu}$ in any direction. We then refine the remaining windows to incorporate $\nu^*$, as shown in \cref{eq:eq2}:
\begin{equation}
    \boldsymbol{\mathcal{C}} = \underset{i \in \boldsymbol{\mathcal{R}}}{\texttt{concat}} (\underset{j \in \mathcal{S}_i}{\texttt{concat}} (\texttt{merge}\left( \nu^*_{i,j}, \mathcal{R}_{i} \right)))
    \label{eq:eq2}
\end{equation}
where $\mathcal{S}_i$ contains indices of surrounding windows that include at least one visual concept for the $i$-th region proposal. These surrounding windows are referred to as \textit{concept windows}. Finally, we sample a \textit{bag of concepts} by merging region proposals with their corresponding sampled concept windows. More implementation details are in \cref{sec:sup:s1}.

\begin{figure}[t]
  \centering
   \includegraphics[width=\linewidth]{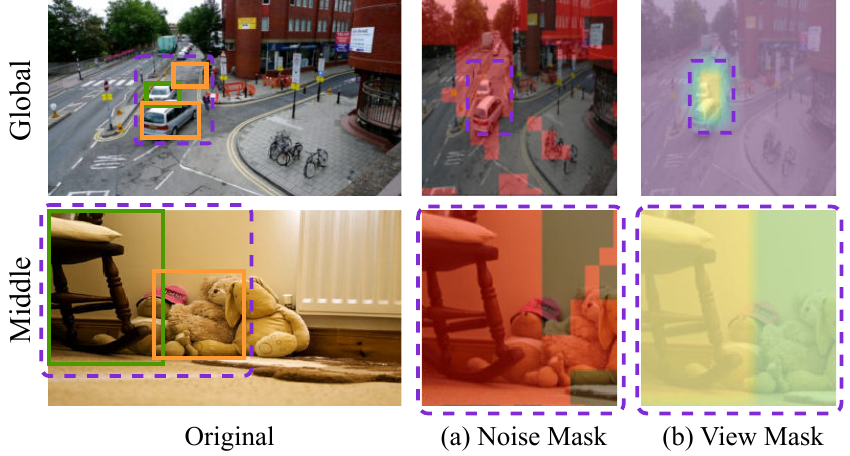}

   \caption{\textbf{Hierarchical views with noise and view masks}. \textbf{(a)} Red masks restrict CLIP attention exclusively to the patch areas. \textbf{(b)} Views become brighter as the weight applied increases, transitioning from blue to yellow.}
   \label{fig:fig3}
\end{figure}

\vspace{-1mm}
\subsection{Aligning Bag of Views}
\label{sec:method:m3}
In this section, we adjust the scale of concepts within a bag to effectively capture surrounding objects of different sizes and their relationships, which BARON cannot achieve due to its fixed window size. To achieve this, we propose a novel representation-switching strategy to explore the optimal scale of concepts within the bag (\cref{sec:method:m3:vs}). For more effective embedding alignment, we utilize previously extracted noise embeddings to generate \textit{noise masks} and different viewpoints within a bag to generate \textit{view masks}. These masks are used to extract the bag-of-view embeddings from the CLIP image encoder (\cref{sec:method:m3:vwa}). We then align these embeddings with their corresponding textual representations (\cref{fig:main}-b).

\vspace{-1mm}
\subsubsection{Representation Switching}
\label{sec:method:m3:vs}    
To effectively learn embeddings for objects of varying sizes, we introduce a novel technique for switching representations across regions, concepts, and \textit{views}. Note that concepts within a bag are formed by merging region proposals with either corresponding neighboring regions by BARON or concept windows. Herein, regions represent potential individual objects identified by BARON, while concepts extend regions by capturing contextual relationships. We introduce a new representation called views, which refine the previous representations by optimizing their scale.

To determine the views, we establish three hierarchical view levels (\eg global, middle, and local). As shown in \cref{fig:main}-b, we define each concept within the bag as the local view, the entire image as the global view, and the merged crop of all concepts as the middle view. We then select a representative view $\mathcal{V}$ for each concept based on differences in view size and the number of visual concepts among the candidate views. Specifically, we compare the local view with its parent view in a greedy fashion (\eg global $\rightarrow$ middle $\rightarrow$ local). In each comparison, we switch the target representation to the current parent view if it exceeds the threshold $\tau$. The threshold is calculated as $\tau = r \times \frac{|L - P|}{P}$; where $r$ is the local-to-parent view size ratio, and $L$ and $P$ represent the number of $\boldsymbol{\nu}$ in the local and parent view, respectively. When the local view closely matches the parent view in size, we increase the probability of switching to the parent. This carries a relatively low risk of complicating the scene. Conversely, we reduce the probability when the parent view contains relatively few co-occurring visual concepts compared to the local view. This representation-switching strategy leads to an optimal scale for each concept within the bag, resulting in a \textit{bag of views}.

\begin{table*}[ht]
  \begin{minipage}{0.72\linewidth}
      \caption{Comparison with previous state-of-the-art methods on the OV-COCO and OV-LVIS benchmarks for open-vocabulary object detection. The symbol $\dagger$ denotes the re-implemented ViLD model.}
      \label{tab:main}
      \vspace{-2.5mm}
      \centering
      \resizebox{\textwidth}{!}{ 
      \begin{tabular}{cc}
          \begin{minipage}[t]{0.55\linewidth} 
              \centering
              \caption*{(a) OV-COCO benchmark}
              \vspace{-2mm}
              \begin{tabular}{l|c|c}
                  \hline
                  Methods & Backbone & $\text{AP}^{\text{novel}}_{50}$ \\ \hline
                  OVR-CNN \cite{OVR-CNN} & ViT-B/32 & 22.8 \\ 
                  RegionCLIP \cite{RegionCLIP} & RN50 & 26.8 \\ 
                  ViLD \cite{vild} & ViT-B/32 & 27.6 \\ 
                  F-VLM \cite{F-VLM} & RN50 & 28.0 \\ 
                  OV-DETR \cite{OV-DETR} & ViT-B/32 & 29.4 \\
                  Detic \cite{detic} & RN50 & 30.3 \\ 
                  PB-OVD \cite{PB-OVD} & RN50 & 30.8 \\ 
                  BARON \cite{BARON} & RN50 & 34.0 \\
                  SAMP \cite{SAMP} & RN50 & 34.8 \\ 
                  BIND \cite{bind} & RN50 & 36.3 \\ 
                  \rowcolor{blue!20} \method (Ours) & RN50 & \textbf{36.6} \\ \hline
              \end{tabular}
          \end{minipage} 
          & 
          \begin{minipage}[t]{0.55\linewidth} 
              \centering
              \caption*{(b) OV-LVIS benchmark}
              \vspace{-2mm}
              \begin{tabular}{l|c|c}
                  \hline
                  Methods & Backbone & mAP$^{\text{mask}}_{r}$ \\ \hline
                  ViLD \cite{vild} & ViT-B/32 & 16.6 \\ 
                  ViLD$^{\dagger}$ & ViT-B/32 & 16.8 \\
                  RegionCLIP \cite{RegionCLIP} & RN50 & 17.1 \\ 
                  OV-DETR \cite{OV-DETR} & ViT-B/32 & 17.4 \\ 
                  Detic \cite{detic} & RN50 & 19.5 \\ 
                  DetPro \cite{detpro} & RN50 & 19.8 \\ 
                  OC-OVD \cite{OC-OVD} & RN50 & 21.1 \\ 
                  OADP \cite{OADP} & ViT-B/32 & 21.7 \\ 
                  VLDet \cite{VLDet} & RN50 & 21.7 \\
                  BARON \cite{BARON} & RN50 & 22.6 \\
                  \rowcolor{blue!20} \method (Ours) & RN50 & \textbf{23.1} \\ \hline
              \end{tabular}
          \end{minipage} 
      \end{tabular}
      }

      \vspace{2.5mm}
      \centering
        \caption{Effectiveness of main components of \method. PFLOPs stands for PetaFLOPs, which represents the total computational cost of the CLIP modules (MLP, CNN, and attention) during training.}
        \label{tab:t2}
        \vspace{-2mm}
        \resizebox{\textwidth}{!}{%
        \begin{tabular}{ccccccc}
            \toprule
            Method & Adaptive Sampling & Noise Mask & View Mask & PFLOPs & $\text{AP}^{\text{novel}}_{50}$ & \textcolor{gray}{$\text{AP}^{\text{base}}_{50}$} \\
            \midrule
            BARON & - & - & - & 55.3 & 34.0 & \textcolor{gray}{60.4}  \\
            \method (Ours) & - & \cmark & - & 25.7 & 34.2 & \textcolor{gray}{55.6} \\
            & \cmark & \cmark & - & 11.5 & 36.0 & \textcolor{gray}{58.0} \\
            \rowcolor{blue!20} & \cmark & \cmark & \cmark & \textbf{10.9} & \textbf{36.6} & \textcolor{gray}{58.9} \\
            \bottomrule
        \end{tabular}
        }
  \end{minipage}%
  \hfill
  \begin{minipage}{0.24\linewidth}
      \raggedleft 
      \caption{Sampling distance $\eta$.}
      \label{tab:t3}
      \vspace{-2mm}
      \centering
      \resizebox{0.8\linewidth}{!}{
      \begin{tabular}{ccc}
          \toprule
          $\eta$ & $\text{AP}^{\text{novel}}_{50}$ & \textcolor{gray}{$\text{AP}^{\text{base}}_{50}$} \\
          \midrule
          0.2 & 36.4 & \textcolor{gray}{57.6} \\ 
          \rowcolor{blue!20} 0.4 & \textbf{36.6} & \textcolor{gray}{58.9} \\ 
          0.8 & 35.8 & \textcolor{gray}{58.1} \\ 
          \bottomrule
      \end{tabular}
      }
      \vspace{3.5mm}
      \caption{Number of extra region proposals $N$.}
      \label{tab:t4}
      \vspace{-2mm}
      \centering
      \resizebox{0.8\linewidth}{!}{
      \begin{tabular}{ccc}
          \toprule
          $N$ & $\text{AP}^{\text{novel}}_{50}$ & \textcolor{gray}{$\text{AP}^{\text{base}}_{50}$} \\
          \midrule
          0 & 35.3 & \textcolor{gray}{57.0} \\ 
          2 & 36.1 & \textcolor{gray}{58.4} \\ 
          \rowcolor{blue!20} 3 & \textbf{36.6} & \textcolor{gray}{58.9} \\ 
          4 & 36.3 & \textcolor{gray}{57.2} \\ 
          \bottomrule        
      \end{tabular}
      }
      
      \vspace{3.5mm}
      \centering
      \caption{Noise mask threshold $\tau$. $\mu$ and $\sigma$ are driven from the distribution $\mathcal{I}(\boldsymbol{\mathcal{V}}) \otimes \chi$.}
      \label{tab:t5}
      \vspace{-2mm}
      \resizebox{\linewidth}{!}{
      \begin{tabular}{ccc}
          \hline
          $\tau$ & $\text{AP}^{\text{novel}}_{50}$ & \textcolor{gray}{$\text{AP}^{\text{base}}_{50}$} \\ \hline
          $\mu+2\sigma$ & 35.4 & \textcolor{gray}{55.2} \\
          \rowcolor{blue!20} $\mu+4\sigma$ & \textbf{36.6} & \textcolor{gray}{58.9} \\
          $\mu+8\sigma$ & 36.2 & \textcolor{gray}{60.1} \\ \hline
      \end{tabular}
      }
  \end{minipage}
\end{table*}

\vspace{-1.5mm}
\subsubsection{View-wise Attention}
\label{sec:method:m3:vwa}
For more effective view embedding alignment, we introduce two key improvements: 1) diminishing the influence of noise (\eg background), and 2) understanding the importance of each view among the sampled views within the bag, where representation switching occurs.

For (1), we design noise masks $\boldsymbol{\mathcal{N}}$ to reduce the influence of noisy patches when extracting embeddings from CLIP. To achieve this, we utilize previously extracted noise embeddings $\chi$ which typically represent general backgrounds. We then apply an attention mask to patches with similarity to the noise embeddings $\chi$ above $\tau$, as shown in \cref{eq:eq3}:
\begin{equation}
    \boldsymbol{\mathcal{N}} = 
        \begin{cases}
        -\infty, & \text{if } \mathcal{I}(\boldsymbol{\mathcal{V}}) \otimes \chi > \tau, \\
        0, & \text{otherwise},
    \end{cases}
\label{eq:eq3}
\end{equation}
where $\mathcal{I}$ is the image encoder and $\otimes$ represents CLIP similarity. We validate the optimal threshold search with visualizations at different thresholds in \cref{sec:sup:s2:mask}.

For (2), we devise view masks $\boldsymbol{\mathcal{M}}$ to enhance the representation based on the importance of each view. Using hyperparameters that define the importance of each view, we empirically analyze the importance of patches corresponding to each view. Herein, the view masks store the importance of each pixel for the respective view. These masks are then multiplied by the softmax output in CLIP's attention mechanism to adjust the influence of patches corresponding to each view. In \cref{fig:fig3}-b, the view masks reflect view importance, with yellow areas indicating higher importance. As shown in \cref{fig:main}-b, we use both masks to extract the view embedding $\boldsymbol{\mathcal{F}}$ from CLIP, as shown in \cref{eq:eq4}:
\begin{equation}
    \boldsymbol{\mathcal{F}} = \left(\texttt{softmax}\left(\frac{\textbf{QK}^\top}{\sqrt{d}} + \boldsymbol{\mathcal{N}}\right) \odot \boldsymbol{\mathcal{M}} \right) \times \textbf{V},
\label{eq:eq4}
\end{equation}%
where $d$ is the CLIP dimension, $\odot$ denotes element-wise multiplication, and $\textbf{Q}$, $\textbf{K}$, and $\textbf{V}$ are the queries, keys, and values of the views within the bag.
\section{Experiments}
\label{sec:exp}

\noindent \textbf{Datasets.}
We evaluate our method using two well-known OVD datasets: OV-COCO \cite{coco} and OV-LVIS \cite{lvis}. For the OV-COCO dataset, we adopt the category split approach from OVR-CNN \cite{OVR-CNN}, dividing the object categories into 48 base categories and 17 novel categories. For the OV-LVIS dataset, we follow ViLD \cite{vild}, separating the 337 rare categories into novel categories and grouping the remaining common and frequent categories into base categories.

\noindent \textbf{Evaluation Metrics.}
We assess detection performance on both base and novel categories.
For OV-COCO, we follow OVR-CNN \cite{OVR-CNN} report the box AP at an IoU threshold of 0.5, denoted as AP50. For OV-LVIS, we report the mean Average Precision (mAP) of masks averaged on IoUs from 0.5 to 0.95. The primary metrics for evaluating open-vocabulary detection performance are the $\text{AP}_{50}$ of novel categories ($\text{AP}^{\text{novel}}_{50}$) for OV-COCO and the mAP of rare categories ($\text{AP}_r$) for OV-LVIS.

\noindent \textbf{Implementation Details.}
We build our method on Faster R-CNN \cite{fasterrcnn} with ResNet50-FPN \cite{fpn}. For a fair comparison, we initialize the backbone network with weights pre-trained by SOCO \cite{soco} and use synchronized Batch Normalization (SyncBN) \cite{syncbn}, as done in recent studies \cite{BARON, detpro}. For the main experiments on OV-COCO and OV-LVIS \cite{coco,lvis}, we choose the $1\times$ and $2\times$ schedules. We use the CLIP model \cite{CLIP} based on ViT-B-16 \cite{vit} as our pre-trained visual language model (VLM). For category name prompts, we default to the hand-crafted prompts from ViLD \cite{vild} in all our experiments on OV-COCO. We use learned prompts following DetPro for experiments on OV-LVIS.

\subsection{Main Results}
\noindent \textbf{OV-COCO.} We compare previous state-of-the-art methods on the OV-COCO (\cref{tab:main}-a). While OV-DETR \cite{OV-DETR}, which builds on Deformable DETR \cite{Deformable-DETR}, outperforms Faster R-CNN \cite{fasterrcnn} on base categories, \method surpasses OV-DETR by 7.2 $\text{AP}_{50}$ on OV-COCO for novel categories. \method even outperforms recent studies \cite{detpro, PB-OVD} that use sophisticated pseudo-labeling. Furthermore, \method outperforms BARON \cite{BARON} by 2.6 $\text{AP}_{50}$ for novel categories, despite the more extensive use of CLIP resources by BARON. We show that \method achieves the best performance across all scenarios.

\noindent \textbf{OV-LVIS.} We compare \method with previous state-of-the-art methods on the OV-LVIS benchmark (\cref{tab:main}-b). Since OV-LVIS has more detailed annotations, it is generally a more challenging benchmark. As a result, the performance observed on OV-LVIS tends to be smaller than OV-COCO. Nevertheless, \method surpasses OV-DETR \cite{OV-DETR} by 5.7 on OV-LVIS for novel categories even in different backbones (\eg Vision Transformer \cite{vit}). \method also performs better than the caption-based models \cite{PB-OVD, RegionCLIP} that use additional caption data. \method also outperforms BARON by 0.5 $\text{AP}_r$ on OV-LVIS. We highlight that \method achieves superior performance in all situations.

However, with the hyperparameters set for OV-COCO, \method shows relatively smaller improvements on benchmarks like OV-LVIS, which contain finer annotations. We believe that this setup is optimized for the coarser annotations of OV-COCO. We provide guidance on hyperparameter tuning for our method on the OVD datasets in \cref{sec:sup:config}.

\subsection{Ablation Study}
In this section, we ablate the effectiveness of components in our method on the OV-COCO benchmark.

\noindent \textbf{Effectiveness of Bag of Views.}
We evaluate the effectiveness of each component of our method (\cref{tab:t2}). First, we evaluate the effect of the noise mask independently. It is applied to each region within a bag, which is obtained by BARON's sampler and used as input for CLIP. We demonstrate that the mask effectively reduces CLIP's computational costs by more than half without sacrificing performance on novel categories. This indicates that the noise mask successfully removes unnecessary background patches from the CLIP's attention mechanism.

Furthermore, by employing our adaptive sampling, our approach already outperforms the baseline \cite{BARON} by 2.6 AP$^{\text{novel}}_{50}$ while significantly reducing computational costs by 79.2\%. This demonstrates that our sampler leverages the generalization power of VLMs more effectively than BARON. It is possible by aligning co-occurring concepts around region proposals and discarding unimportant ones.

Lastly, when we apply the view mask along with representation switching, we find that our method further improves performance by 0.6 AP$^{\text{novel}}_{50}$ and increases computational efficiency by reducing 0.6 PFLOPs.

\noindent \textbf{Sampling Distance.}
We analyze the impact of the sampling distance $\eta$ on the candidate visual concepts $\boldsymbol{\nu}$. As shown in \cref{tab:t3}, we observe that a larger concept window with a high $\eta$ may capture a more complex scene with excessive objects and noise, leading to performance degradation. Nevertheless, our method ensures exclusive learning of concepts within a bag by trimming overlap with other region proposals, resulting in a performance boost even with $\eta$ set to 0.8.

\noindent \textbf{Number of Extra Region Proposals.}
We study the impact of the number of extra region proposals $N$, inspired by Faster R-CNN's tendency to predict a few region proposals per image after NMS. With $N$ set to 0 by default, our method achieves results comparable to BARON, with a 1.3 AP$^{\text{novel}}_{50}$ increase. As $N$ increases, our method boosts performance linearly by identifying more key visual concepts through additional edges from varying directions within each image. However, when $N$ exceeds a certain threshold, the number of these concepts becomes saturated, yielding no additional performance gains.

\begin{figure*}[t]
  \centering
   \includegraphics[width=\linewidth]{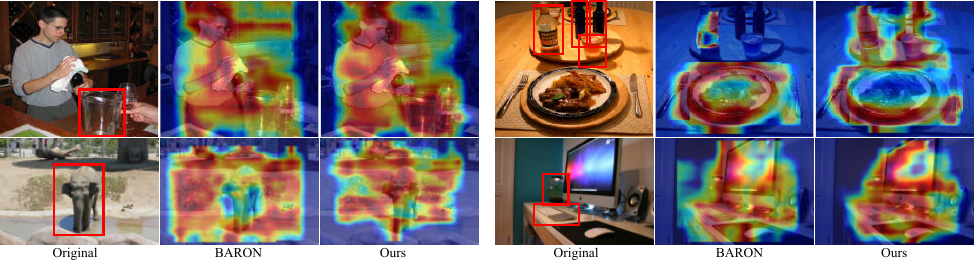}
   \vspace{-6mm}
   \caption{\textbf{Qualitative comparisons between \method and BARON}. The images are from COCO’s validation set. Red boxes in the original images highlight the novel categories. The feature maps respond to objects identified by the detector. The novel categories are `cup’, `keyboard,' `elephant,' and `wine glass’. \method detects novel objects missed by the baseline.}
   \label{fig:qual}
\end{figure*}

\begin{figure*}[t]
  \centering
   \includegraphics[width=\linewidth]{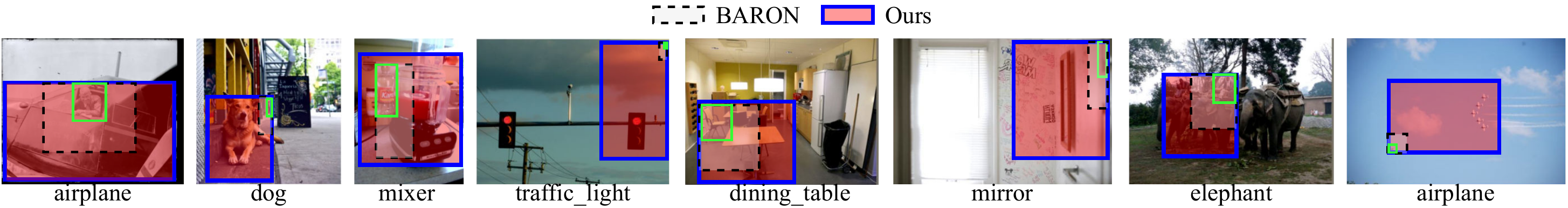}

   \caption{\textbf{Visualization between \method and BARON}. Green boxes indicate base categories; each image has a label representing its corresponding novel category, such as `dog,' `airplane' and `mirror'.}
   \label{fig:qual2}
\end{figure*}

\noindent \textbf{Noise Mask Threshold.}
We ablate the noise mask $\boldsymbol{\mathcal{N}}$ using different thresholds from the distribution $\mathcal{I}(\boldsymbol{\mathcal{V}}) \otimes \chi$ in \cref{tab:t5}. $\tau$ is set to combinations of $\mu$ and $\sigma$ from the distribution. We observe the noise mask becomes stricter in removing unnecessary background areas as $\tau$ increases. The best result on novel categories is achieved with $\tau = \mu+4\sigma$. For further analysis in \cref{sec:sup:s2:mask}, we present a visualization showing the effectiveness of this mask in capturing key concepts and minimizing unnecessary areas across different thresholds.

\begin{table}[t]
    \centering
    \caption{Exploring the effects of views on the COCO training set.}
    \resizebox{0.67\linewidth}{!}{
    \begin{tabular}{ccccc}
        \toprule
        $\delta^{\text{global}}$ & $\delta^{\text{middle}}$ & $\delta^{\text{local}}$ & $\text{AP}^{\text{novel}}_{50}$ & \textcolor{gray}{$\text{AP}^{\text{base}}_{50}$} \\
        \midrule
        1.0 & - & - & 16.9 & \textcolor{gray}{47.7} \\
        - & 1.0 & - & 30.8 & \textcolor{gray}{53.8} \\
        - & - & 1.0 & 36.0 & \textcolor{gray}{58.0} \\ \hdashline
        0.2 & 0.8 & 1.0 & 31.7 & \textcolor{gray}{54.2} \\
        0.1 & 0.8 & 1.0 & 33.9 & \textcolor{gray}{56.8} \\ 
        0.0 & 0.6 & 1.0 & 35.6 & \textcolor{gray}{57.2} \\
        \rowcolor{blue!20} 0.0 & 0.8 & 1.0 & \textbf{36.6} & \textcolor{gray}{58.9} \\
        0.0 & 1.0 & 1.0 & 36.3 & \textcolor{gray}{59.4} \\
        \bottomrule        
    \end{tabular}
    }
    \label{tab:gml}
    \vspace{-3mm}
\end{table}

\begin{table}[t]
    \centering
    \caption{Exploring the effects of different sampling strategies.}
    \resizebox{0.8\linewidth}{!}{
    \begin{tabular}{cccc}
        \toprule
        Sampling strategy & $\text{AP}^{\text{novel}}_{50}$ & \textcolor{gray}{$\text{AP}^{\text{base}}_{50}$} & \#regions \\
        \midrule
        Grid & 25.4 & \textcolor{gray}{58.0} & 36 \\ 
        Random & 27.3 & \textcolor{gray}{53.3} & 36 \\ 
        Random-Tight & 29.5 & \textcolor{gray}{56.9} & 36 \\ 
        Random-Neighbor & 30.7 & \textcolor{gray}{56.9} & 36 \\ 
        BARON (reduced) & 32.2 & \textcolor{gray}{58.3} & 36 \\
        BARON & 34.0 & \textcolor{gray}{60.4} & 216 \\ 
        \rowcolor{blue!20} \method (Ours) & \textbf{36.6} & \textcolor{gray}{58.9} & 56 \\
        \bottomrule        
    \end{tabular}
    }
    \label{tab:t7}
    \vspace{-3mm}
\end{table}

\noindent \textbf{View Importance.}
We study the impact of each view on the performance for novel categories by adjusting combinations of view weights $\delta^{\text{global, middle, local}}$ during training. We first isolate each view by setting its weight to 1.0 while neglecting the others, without triggering representation switching. In \cref{tab:gml}, forcing the global view leads to a performance drop to 16.9 AP$^{\text{novel}}_{50}$, while forcing the local view maintains 36.0 AP$^{\text{novel}}_{50}$. This linear performance drop suggests that the global view captures overly complex scenes.

Furthermore, within each view, we set $\delta^{\text{local}}$ to 1.0, as the local view is more likely to contain the ground truth target. In \cref{tab:gml}, we observe a significant performance drop as the weight of $\delta^{\text{global}}$ increases by 0.1. The best result on novel categories is achieved by setting $\delta^{\text{global}}$ to 0.0, $\delta^{\text{middle}}$ to 0.8, and $\delta^{\text{local}}$ to 1.0. We leave the study of leveraging the entire image to enhance performance as future work.

\noindent \textbf{Sampling Strategy.}
We compare our adaptive sampling with other strategies (\cref{tab:t7}). The grid sampling strategy divides the image into equal grids, similar to the pretraining stage in OVR-CNN \cite{OVR-CNN}. The random sampling strategy arbitrarily selects region proposals to form a bag of regions, usually representing the entire image. The Random-Tight strategy focuses on a cropped area that tightly encloses the selected regions instead of the entire image. The Random-Neighbor strategy selects two nearby regions with GIoU $> 0.5$ for each central region, ensuring 36 regions from 12 centers. The BARON (reduced) limits the number of region proposals per image to 12, taking one bag per proposal. We show our method outperforms all these sampling strategies.

\subsection{Qualitative Research}
In this section, we present qualitative results to further analyze the effectiveness of our method.

\noindent \textbf{Qualitative results.}
We visualize the predictions of detectors learned through our method and BARON (\cref{fig:qual}). We visualize the feature map response to both base and novel categories using Eigen-CAM \cite{eigencam}. We find that our method produces discriminative responses at novel category locations, while BARON shows weaker or scattered responses. For example, our method accurately attends to novel categories like `elephant,' which BARON often misses.

\noindent \textbf{Visualization of sampling results.}
We compare the sampling results of our method and BARON on the two OVD datasets. Each image shows labels for novel categories, with base categories in green boxes (\cref{fig:qual2}). BARON, which samples neighboring regions based on a fixed window size of the region proposal, often produces a bag of regions that do not fully capture semantic concepts, such as novel objects. In contrast, our method samples a bag of views based on surrounding meaningful concepts, capturing novel objects more effectively than BARON. Thus, we demonstrate that our method fosters a more powerful and efficient compositional structure within VLMs.
\vspace{-1mm}
\section{Conclusion}
\label{sec:conclusion}

This paper explores an improved compositional structure of semantic concepts through a bag of views in OVD. The compositional structure is mainly about the co-occurrence of objects. This leverages large-scale VLMs' ability to represent the compositional structure of concepts within image-text pairs. We develop an adaptive sampling strategy to group contextually related concepts into a bag. Concepts within a bag are then adjusted to views considered optimal scales, with masks applied to enhance embedding alignment. Finally, we adopt a distillation-based approach to align the detector's bag-of-view representations with those of pre-trained VLMs. Our method outperforms previous state-of-the-art methods on the OVD benchmarks while significantly enhancing VLM utilization efficiency.
{
    
    \small
    \bibliographystyle{ieeenat_fullname}
    \bibliography{main}
}

\addtocontents{toc}{\protect\setcounter{tocdepth}{2}}

\clearpage
\setcounter{page}{1}
\maketitlesupplementary

\renewcommand{\thesection}{\Alph{section}}
\setcounter{section}{0}

\tableofcontents
\newpage




\section{Implementation Details}
\subsection{Implementation}
\label{sec:sup:s1}
\noindent \textbf{Computing unnecessary neighboring regions.}
We outline the process for calculating the ratio of unnecessary neighboring regions extracted by BARON~\cite{BARON}. In our analysis, these neighboring regions are considered unnecessary if their overlap with the ground-truth (GT) boxes is less than 85\% IoU or if their embeddings have a cosine similarity score above 0.8 compared to noise embeddings.

Our threshold selection is inspired by the non-maximum suppression (NMS) process in Faster R-CNN~\cite{fasterrcnn}. Faster R-CNN uses NMS to filter out overlapping region proposals with lower objectness scores, retaining only a few proposals. Specifically, two RPN boxes are considered overlapping when their IoU is 0.1, allowing the detector to focus on the most likely RPN box within the surrounding area. For consistency, we adhere to BARON's hyperparameter settings when applying Faster R-CNN. This forms the basis for our hyperparameter choices. Specifically, we classify regions with an IoU approximately close to $1 - 0.1 = 0.9 \pm 0.05$ as sufficiently representative of a GT box, while regions with a lower IoU are deemed to ambiguously represent the GT.

Additionally, we include a feature-level comparison using cosine similarity between noise embeddings and the regional embeddings of RPN boxes. The learnable noise embeddings are generated by training an additional linear layer on the model’s class tokens. Due to the noisy nature of similarity distributions, we only consider background regions with a cosine similarity of approximately 0.8 or higher as unnecessary. Consequently, the hyperparameters for IoU and the similarity threshold are set to 0.85 and 0.8, respectively.

\begin{algorithm*}[h]
\caption{Aligning Bag of Views}\label{alg:alg1}
\begin{algorithmic}[1]
\Require an input image $\mathcal{I}$, region proposals $\boldsymbol{\mathcal{R}}$, top-$k$ proposals $\boldsymbol{\mathcal{R}}_{\text{top}k}$, noise embeddings $\chi$
\Hyperparams interval $\Delta$, number of edges $E$, number of extra region proposals $N$, scaling factor for noise mask  $s$, distance threshold $\eta$, weight for distance $\lambda$, view weights $\delta^{\{\text{global, middle, local}\}}$
\Ensure CLIP features $\pmb{\mathcal{F}}$

\State \textbf{Step 1: Forming Bag of Concepts}
\State $\boldsymbol{\mathcal{R}}_{\text{reduced}},\ \boldsymbol{\mathcal{R}}_{\text{added}} \gets \text{preprocess}(\boldsymbol{\mathcal{R}}, \boldsymbol{\mathcal{R}}_{\text{top}k})$ \Comment{\cref{eq:sup:nonoverlap}, \cref{eq:sup:added} and \cref{eq:sup:score}}
\State $\mathcal{G} \gets \text{generate-canvas}(\mathcal{G}, \boldsymbol{\mathcal{R}}_{\text{reduced}})$
\For{\textbf{each distinct pair} $(b_{\text{start}}, b_{\text{end}}) \in \boldsymbol{\mathcal{R}}_{\text{added}}$ \textbf{where} $b_{\text{start}} \neq b_{\text{end}}$}
    \State $\boldsymbol{\mathcal{E}} \gets \text{generate-edges}(b_{\text{start}}, b_{\text{end}}, \mathcal{G}, E)$
    \State $\boldsymbol{\nu} \gets \texttt{concat}(\boldsymbol{\nu}, \text{extract-concepts}(\boldsymbol{\mathcal{R}}_{\text{reduced}}, \boldsymbol{\mathcal{E}})$) \Comment{\cref{eq:eq1}}
\EndFor
\State $\boldsymbol{\mathcal{D}} \gets \text{update-concept-windows}(\boldsymbol{\mathcal{R}}, \boldsymbol{\nu}, \eta, \lambda)$ \Comment{\cref{eq:eq2}}

\State \textbf{Step 2: Aligning Bag of Views}
\State $\boldsymbol{\mathcal{R}}^{\text{global}}, \boldsymbol{\mathcal{R}}^{\text{middle}}, \boldsymbol{\mathcal{R}}^{\text{local}}, \boldsymbol{\mathcal{V}} \gets \text{representation-switching}(\boldsymbol{\mathcal{D}}$)
\State $\boldsymbol{\mathcal{M}} \gets \text{compute-view-masks}(\boldsymbol{\mathcal{M}}, \boldsymbol{\mathcal{R}}^{\text{global}}, \boldsymbol{\mathcal{R}}^{\text{middle}}, \boldsymbol{\mathcal{R}}^{\text{local}}$)
\State $\boldsymbol{\mathcal{N}} \gets \text{compute-noise-masks}(\boldsymbol{\mathcal{N}}, \boldsymbol{\mathcal{V}}, \chi$, s) \Comment{\cref{eq:eq3}}
\State $\pmb{\mathcal{F}} \gets \text{extract-clip-features}(\boldsymbol{\mathcal{V}}, \boldsymbol{\mathcal{M}}, \boldsymbol{\mathcal{N}}$) \Comment{\cref{eq:eq4}}

\State \Return $\pmb{\mathcal{F}}$
\end{algorithmic}
\end{algorithm*}

\noindent \textbf{Sampling bag of concepts.}
In this paper, we introduce \textit{concept windows}, which effectively capture key concepts near each region proposal. Open vocabulary object detection (OVD) aims to predict novel objects often unseen during training, emphasizing the need to prevent overfitting to training data. For implementation details, we outline two operations within our adaptive sampling strategy designed to minimize overfitting to easily identifiable objects frequently found in region proposals during training.

1) We design these concept windows to avoid repeatedly including other regional proposals. To achieve this, we trim the concept windows at the boundaries of all region proposals, preventing training bias and reducing the risk of overfitting certain objects while underpredicting others.

2) Additionally, we incorporate BARON's sampling strategy for region proposals that do not involve concept windows at all, ensuring balanced training counts. This typically occurs when the region proposals lack nearby concepts. As a result, \method enables efficient and mutually exclusive training for each proposal.

\subsection{Pseudocode}
We provide the pseudocode for our method in \cref{alg:alg1}. Note that all necessary requirements, such as the noise embeddings $\chi$, have been preprocessed as outlined in the pseudocode. Our method, \method, includes several hyperparameters; however, we find that only four significantly contribute to performance improvement, while the others are fixed for each OVD benchmark. These hyperparameters are highlighted in blue in \cref{sup:tab:config}. Further details on this hyperparameter configuration can be found in \cref{sec:sup:config}.

\noindent \textbf{Forming Bag of Concepts.}
\method first forms a bag of concepts for each region proposal extracted by Faster R-CNN~\cite{fasterrcnn} within an image. Here, the region proposals are reduced to $\mathcal{R}_{\text{reduced}}$ for more efficient canvas construction, as described in \cref{sec:sup:s1:overlap}. To achieve this, we build a novel structure named \textit{canvas} which aids in effectively searching key concepts between the region proposals. Specifically, the canvas contains coordinates of interval $\Delta$ which direct to areas where RPN boxes with high objectness scores are densely populated. We sample these key concepts from overly abundant and noisy RPN boxes that potentially represent novel objects through edge generation on the canvas. Then, we form a bag of concepts by merging each surrounding window of region proposals with the corresponding key concepts, as shown in \cref{eq:eq2}.

\noindent \textbf{Aligning Bag of Views.}
To enhance the concept embeddings, we introduce representation switching, which identifies the best representation, termed the representative view. This view is chosen from three hierarchical views derived from the bag of concepts, based on the number of co-occurring concepts and the view size ratio. The selected view is further refined using two masks: the view mask and the noise mask. The view mask emphasizes the importance of each view, while the noise mask suppresses noisy representations (\eg backgrounds).



\subsection{Non-overlapping RPN Boxes}
\label{sec:sup:s1:overlap}
We introduce a simple yet effective technique to accelerate the computation of existence probabilities for numerous top-$k$ RPN boxes around each coordinate on the canvas. The key idea is to selectively reduce the top-$k$ RPN boxes. To ensure uniform probability calculations across all image regions, the RPN boxes need to be sparsely distributed. RPN boxes already fulfill this requirement but may demand significant processing time due to their excessive abundance. To address this, we optionally use greedy sampling to select non-overlapping RPN boxes from the top-$k$ boxes in the image, as shown in \cref{eq:sup:nonoverlap}:
\begin{equation}
    \boldsymbol{\mathcal{R}}_{\text{reduced}} = \{ r \in \boldsymbol{\mathcal{R}}_{\text{top}k} \mid \forall r' \in \boldsymbol{\mathcal{R}}, \text{IoU}(r, r') > \tau \}
    \label{eq:sup:nonoverlap}
\end{equation}%
where $\tau$ indicates an IoU threshold default to 0.0, IoU denotes Intersection over Union, $\mathcal{R}$ represents region proposals, and $\mathcal{R}_{\text{top}k}$ refers to the top-$k$ region proposals extracted by Faster R-CNN~\cite{fasterrcnn}.

In this process, we extract RPN boxes that do not overlap with the region proposals and ignore very small RPN boxes for simplicity. As a result, this greedy sampling effectively reduces the top-$k$ RPN boxes, allowing probability calculations to be performed based on a smaller number of RPN boxes, thus accelerating the computation process during the canvas creation. The visualization of the greedy sampling is shown as notable probabilities in \cref{fig:sup_main}-b.

\subsection{$k$-farthest RPN Boxes}
\label{sec:sup:s1:fartest}
In this paper, we increase the number of region proposals to encourage more active edge generation, facilitating more effective concept sampling. Note that the meaningful concepts are RPN boxes that are densely populated and have high objectness scores. This approach addresses the issue observed in Faster R-CNN~\cite{fasterrcnn}, where only a few region proposals are generated per image after NMS. Specifically, we often observe cases where fewer than three small-sized region proposals are generated per image. We believe that this problem may be due to the RPN being overfitted to the training data, leading it to focus primarily on objects belonging to well-defined base categories. Note that we follow BARON's hyperparameter settings for the RPN to ensure a fair comparison.

As shown in \cref{fig:sup_main}-a and b, the number of region proposals extracted by the RPN (highlighted in green) is often fewer than four in randomly sampled images from the OVD datasets. Furthermore, these proposals can correspond to meaningless objects and are of small window sizes. For instance, in the ``mirror'' sample, there are no objects belonging to any category.

To address this issue, we incorporate additional region proposals by augmenting the default region proposals $\boldsymbol{\mathcal{R}}$ with extra RPN boxes from $\boldsymbol{\mathcal{R}}_{\text{reduced}}$, resulting in $\boldsymbol{\mathcal{R}}_{\text{added}}$. Specifically, we sample RPN boxes that are, on average, a distance of $\eta$ away from $\boldsymbol{\mathcal{R}}$, as described in \cref{eq:sup:added}:
\begin{equation}
    \boldsymbol{\mathcal{R}}_{\text{added}} = \left\{ \boldsymbol{r} \in \boldsymbol{\mathcal{R}}_{\text{added}} \; \middle| \; \boldsymbol{r} \in \mathop{\arg\max}^{N}_{r \in \boldsymbol{\mathcal{R}}_{\text{reduced}}} S(r) \right\}
    \label{eq:sup:added}
\end{equation}
where $S(r)$ represents a score function as defined in \cref{eq:sup:score}, $N$ denotes the number of additional region proposals, and $\boldsymbol{r}$ include $\{r_1,\ldots,r_N\}$.
\begin{equation}
    \begin{aligned}
        S(r) &= \lambda \cdot \left\| \frac{1}{|\boldsymbol{\mathcal{R}}|} \sum_{r' \in \boldsymbol{\mathcal{R}}} \min_{p, q \in C(r, r')} \| p - p' \| \right\|_{\eta} \\
        &\quad + (1 - \lambda) \cdot \texttt{objectness}(r)
    \end{aligned}
    \label{eq:sup:score}
\end{equation}
where $\texttt{objectness}(\cdot)$ denotes the objectness score of RPN boxes, $\lambda$ represents the weight assigned to the distance, $\| \cdot \|_{\eta}$ indicates the normalization based on the distance threshold $\eta$, and $C(\cdot)$ returns the center coordinates of the boxes. Note that $r'$ and $r$ are distinct from each other and sampled in a way that avoids duplicate values.

By leveraging this increased number of region proposals, key concepts distributed across the image can be identified from various directions through edges. Consequently, \method demonstrates robustness in detecting key concepts (\eg novel objects) via edge generation, even in samples where base categories are extremely absent.

\subsection{Calculating FLOPs for VLMs}
In this paper, \method utilizes the CLIP model \cite{CLIP} based on ViT-B-16 \cite{vit}. We use Floating Point Operations (FLOPs) to measure the computational load of such VLMs. FLOPs is a widely used metric in machine learning to quantify the computing power of a computer or processor. The operations in the CLIP model are divided into three modules: convolutional neural network (CNN) layer, self-attention, and multilayer perceptron (MLP). We calculate and sum the FLOPs of each component to obtain the total load for a single CLIP inference. The areas where \method results in actual FLOPs changes are in the self-attention and MLP components, where our noise masks $\boldsymbol{\mathcal{N}}$ function as attention masks. As a result, we demonstrate that \method achieves 80.3\% FLOPs reduction compared to BARON.

\begin{figure*}[t]
  \centering
   \includegraphics[width=\linewidth]{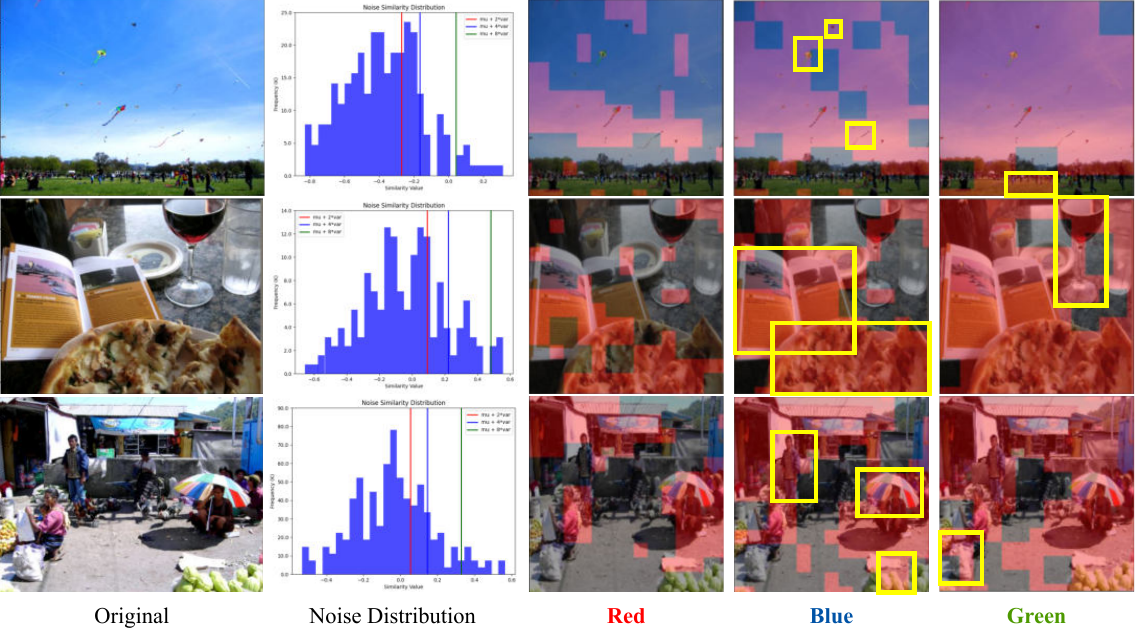}

   \caption{\textbf{Noise mask thresholds}. We obtain the noise similarity distribution by computing the similarity between image features extracted by the CLIP image encoder and noise embeddings $\boldsymbol{\mathcal{N}}$. Red, blue, and green lines represent each threshold described in \cref{tab:t5}, respectively. Yellow boxes in noise masks highlight newly detected objects at each threshold (\eg umbrella, apple, and person). The noise mask filters out redundant areas (\eg background).}
   \label{fig:sup_vis_mask}
\end{figure*}

\noindent \textbf{CNN layer.}
To calculate the FLOPs for a convolutional layer, we first define key parameters: the filter (kernel) size \( K \times K \), the input and output channels \( C_{\text{in}} \) and \( C_{\text{out}} \), and the input size \( H_{\text{in}} \times W_{\text{in}} \). The height and width of the output feature map are determined by the stride, with output dimensions \( H_{\text{out}} = H_{\text{in}} / S_{\text{h}} \) and \( W_{\text{out}} = W_{\text{in}} / S_{\text{w}} \), where \( S_{\text{h}} \) and \( S_{\text{w}} \) are the stride lengths for height and width, respectively. Each convolution operation at a single position requires \( K \times K \times C_{\text{in}} \times C_{\text{out}} \) multiplications and an equal number of additions. Across the entire output feature map, the total FLOPs can be calculated by multiplying the FLOPs per instance by the total number of output positions \( H_{\text{out}} \times W_{\text{out}} \). Thus, the total FLOPs for the convolutional layer is in \cref{eq:sup:cnn}:
\begin{equation}
    \text{FLOPs}_{\text{CNN}} = (K^2C_{\text{in}} C_{\text{out}}) \times (H_{\text{out}}W_{\text{out}})
    \label{eq:sup:cnn}
\end{equation}

\noindent \textbf{Self-attention layer.}
In calculating the FLOPs for the self-attention layer, we define the following key parameters: the attention window size \( W \), embedding dimension \( D \), and number of attention heads \( H \). This is where our noise mask, \( \mathcal{N} \), comes into play. We denote \( \mathcal{N} \) as the attention mask used in self-attention, where it specifically represents the unmasked positions within the window \( W \). This parameter adjusts the computation according to the number of unmasked positions each token can attend to. For a batch of size \( B \), and summing across all \( H \) heads, the total FLOPs for self-attention with the attention mask can be expressed in \cref{eq:sup:sa}:
\begin{equation}
    \text{FLOPs}_{\text{SA}} = 2BHD \times W\mathcal{N}.
    \label{eq:sup:sa}
\end{equation}

\noindent \textbf{MLP layer.}
To compute the FLOPs for the MLP layer, we consider two main fully connected sub-layers: the fully connected layers and the projection layer. For each fully connected layer, the number of FLOPs is calculated by multiplying the input size by the output dimension of the layer. Given a batch size of \( B \) and a window size \( W \), the total FLOPs for the MLP layer is expressed in \cref{eq:sup:mlp}:
\begin{equation}
    \text{FLOPs}_{\text{MLP}} = 2BWD_{\text{fc}} + BWD_{\text{proj}},
    \label{eq:sup:mlp}
\end{equation}
where \( D_{\text{fc}} \) and \( D_{\text{proj}} \) represent the output dimensions of the two sub-layers, respectively. The factor of \( 2 \) accounts for the two multiplications in the first fully connected layer.

\section{Further Analysis}
\label{sec:sup:s2}

\subsection{Noise Masks}
\label{sec:sup:s2:mask}
We find that the noise embeddings $\boldsymbol{\mathcal{N}}$ represent redundant areas (\eg backgrounds). These embeddings are extracted by preemptively training the baseline model with an additional learnable linear layer added to the model's class tokens \cite{BARON}. This allows for proactively learning features for non-category regions. Such design is demonstrated by OVD, which leverages VLMs to indirectly enable the model to learn features for novel objects during training. This enables the noise embeddings to learn to detect regions outside the features of novel objects. Then, we compute the pixel-wise similarities between the noise embeddings and each region embedding during training. We remove redundant patches from CLIP attention that exceed a certain similarity threshold to obtain the noise mask. We find higher similarity scores indicate noisier regions, converging towards background areas.

However, potentially novel objects may still exist in these noisy regions, as the noise embeddings are trained on the training set. This suggests the need to model a certain amount of noise together. In \cref{fig:sup_vis_mask}, we observe that the similarity distribution between the regional embeddings and the noise embeddings varies across all region crops. Note that the mean and variance of the distribution differ. A larger mean indicates that the scene contains mostly noisy representations, while a smaller mean suggests an abundance of primary objects. This implies that fixed similarity thresholds fail to capture both primary objects and appropriate noise within each crop.

To address this issue, we study threshold ablation for each region crop. Specifically, we explore optimal thresholds using combinations of mean and variance from the similarity distribution. The combinations are as follows: ($\mu+2\sigma$), ($\mu+4\sigma$), and ($\mu+8\sigma$); where $\mu$ and $\sigma$ are the mean and variance of the similarity distribution. Note that $\mu$ ensures minimal noise for each crop, while adjusting $\sigma$ with a scaling factor $\alpha$ controls the degree of noise blending. As shown in \cref{fig:sup_vis_mask}, the \textcolor{red}{\textbf{Red}}, \textcolor{blue}{\textbf{Blue}}, and \textcolor{green}{\textbf{Green}} lines within the distribution represent these thresholds, respectively. Then, we visualize the noise masks for each threshold. Each threshold’s color label corresponds to its noise mask. The noise masks contain red patches indicating regions where attention operations occur. The regions often highlight primary objects from either base or novel categories. In contrast, regions without any masks are excluded from CLIP attention operations.

We observe that more patches undergo attention operations as the scaling factor $\sigma$ increases. When $\sigma$ is relatively small (\textcolor{red}{\textbf{Red}}), patches around primary objects are more selectively preserved. However, some objects are often missed if the threshold is too low. This results in unstable mask generation. Conversely, when $\sigma$ is too high (\textcolor{green}{\textbf{Green}}), the model computes CLIP attention to redundant areas, converging toward all patches. This results in unnecessary computational costs for CLIP inference. Instead, we attend to the impact of the \textcolor{blue}{\textbf{Blue}} threshold of $\mu + 4\sigma$. In this case, we observe clear identification of primary objects while effectively removing unnecessary background regions. For example, yellow boxes in noise masks highlight newly detected objects at each threshold in \cref{fig:sup_vis_mask}. At the \textcolor{blue}{\textbf{Blue}} threshold, we find that the number of such boxes is high, while redundant regions are minimized.

\subsection{Hyperparameter Configuration}
\label{sec:sup:config}
\cref{sup:tab:config} details the hyperparameter configuration used in our OV-LVIS and OV-COCO experiments. The configurations for OV-COCO and OV-LVIS are as follows: Both use the SGD optimizer with a momentum of 0.90 and a weight decay of \(2.50 \times 10^{-5}\). The learning rate for OV-COCO is set to 0.04, while for OV-LVIS, it is 0.08. OV-COCO runs for a total of 90K iterations, whereas OV-LVIS runs for 180K iterations. Additionally, the batch size for OV-COCO is 8, while OV-LVIS uses a batch size of 16. To ensure a fair comparison, we adopt the same hyperparameter settings as BARON for constructing a bag, specifically $k$ and $G$.

In this work, \method performs a hyperparameter search involving $\Delta$, $\eta$, $s$, and $\sigma$. Note that the remaining hyperparameters are set to fixed numeric values by default for our experiments. Given that the OV-LVIS benchmark, with its more extensive annotations, is generally considered more challenging than OV-COCO, stricter hyperparameter settings are required for OV-LVIS. However, by applying the same hyperparameter settings from OV-COCO to OV-LVIS, we observe suboptimal performance gains (0.5 increase) on OV-LVIS, as shown in \cref{tab:main}.

To further improve performance, we provide guidelines for adjusting the hyperparameters of our model. Specifically, we believe that the middle view in OV-LVIS can provide as much information as the global view in OV-COCO. This is because the OV-LVIS dataset contains more detailed, fine-grained annotations within a smaller area compared to OV-COCO. Thus, we recommend setting its weight to 0.0 to eliminate its contribution to OV-LVIS. Additionally, we recommend increasing the noise mask threshold, which also serves as a similarity threshold for the noise embeddings. A small threshold may cause the mask to be applied to scattered object areas, leading to the loss of important details. For the OV-LVIS dataset, we also recommend reducing the interval $\Delta$ and the distance threshold $\eta$ to help capture more densely located objects. As a result, these adjustments are aimed at optimizing our model for different settings (\eg benchmarks) and further enhancing its performance.

\begin{table}[t]
    \centering
    \caption{Hyper-parameter configuration of \method. The hyperparameters highlighted in blue contribute relatively more to significant performance gains.}
    \resizebox{\linewidth}{!}{
    \begin{tabular}{ccc}
        \toprule
        \textbf{Configuration} & \textbf{OV-COCO} & \textbf{OV-LVIS} \\
        \midrule
        Optimizer & SGD & SGD \\
        Learning rate & 0.04 & 0.08 \\ 
        Momentum & 0.90 & 0.90 \\
        Weight decay & $2.50 \times 10^{-5}$ & $2.50 \times 10^{-5}$ \\
        Total iterations & 90k & 180k \\ 
        Batch size & 8 & 16 \\ \hdashline 
        Number of region proposals ($k$) & 300 & 500 \\
        Number of bags ($G$) & 3 & 4 \\
        Number of edges ($E$) & 2 & 4 \\ 
        Number of extra region proposals ($N$) & 3 & 5 \\
        Weight for Distance ($\lambda$) & 0.5 & 0.5 \\
        Weight for Aspect ratio ($\alpha$) & 0.5 & 0.5 \\
        \rowcolor{blue!20} interval ($\Delta$) & 100 & 100 \\ 
        \rowcolor{blue!20} Scaling factor for noise mask ($s$) & 4 & 4 \\
        \rowcolor{blue!20} Distance threshold ($\eta$) & 0.4 & 0.4 \\
        \rowcolor{blue!20} View weights ($\delta^{\{\text{global, middle, local}\}})$ & \{-, 0.8, 1.0\} & \{-, 0.8, 1.0\} \\ 
        \bottomrule        
    \end{tabular}
    }
    \label{sup:tab:config}
\end{table}

\subsection{Visualization}
\label{sec:sup:s2:main}
\begin{figure*}[t]
  \centering
   \includegraphics[width=\textwidth, height=\textheight, keepaspectratio]{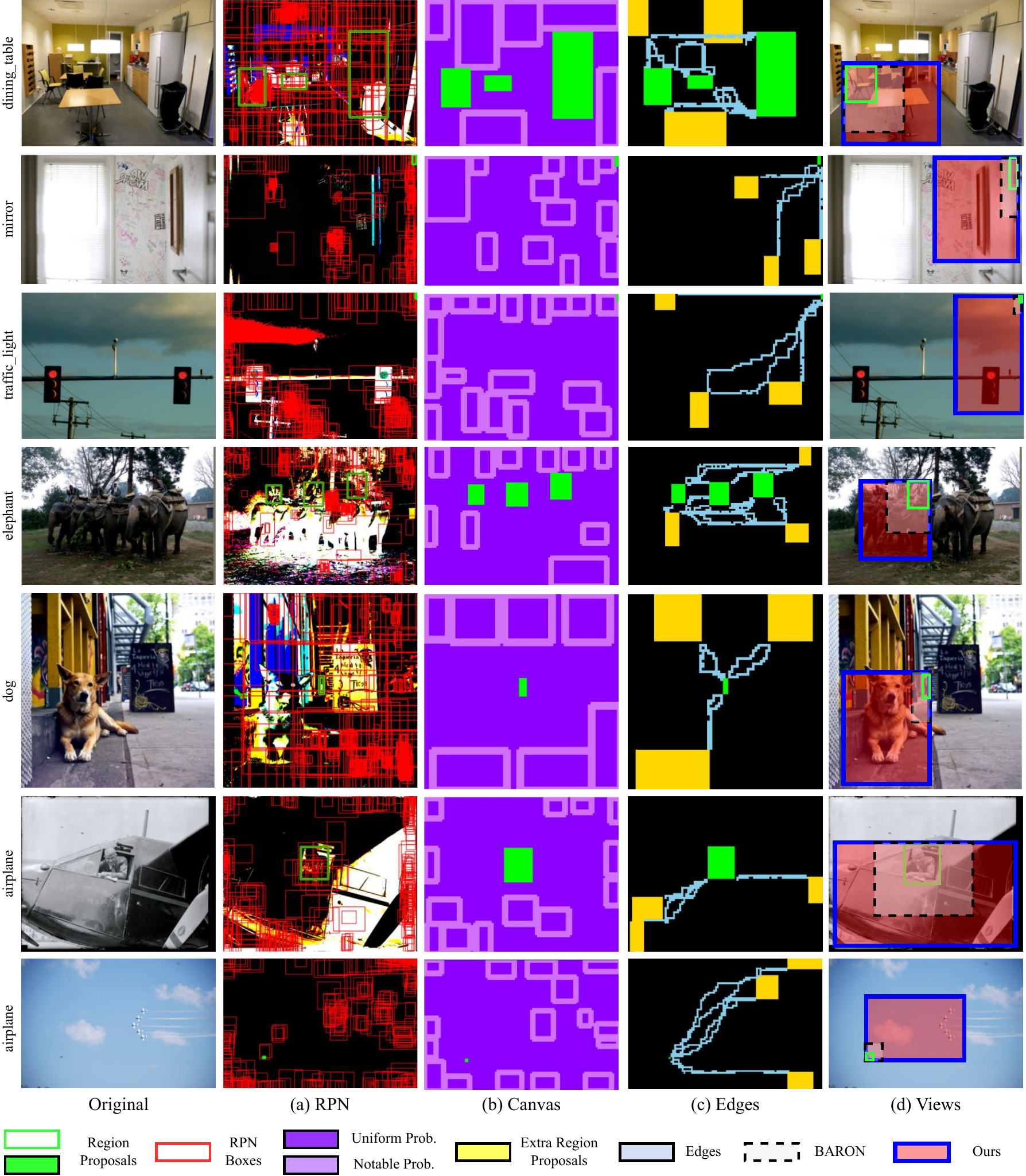}
   \caption{\textbf{Visualization of \method compared to BARON on both the OV-COCO and OV-LVIS datasets}. This figure includes the original image, RPN results, canvas, edges, and views extracted by \method. We demonstrate that \method generates views that more accurately represent surrounding semantic concepts, including novel categories in each image, compared to BARON.}
   \label{fig:sup_main}
\end{figure*}

In this section, we visualize each module of \method in \cref{fig:sup_main}. We aim to demonstrate its effectiveness in capturing meaningful concepts around each region proposal.
We visualize eight images containing novel categories from the OVD datasets. These novel categories (\eg ``dog'', ``mirror'', ``elephant'', and ``traffic\_light'') are listed on the left side of each row. Note that the legend for each box is specified at the bottom of the figure.

First, we analyze the characteristics of the Region Proposal Network (RPN), which has the potential to detect novel objects~\cite{fasterrcnn}. We then examine its results and limitations in capturing key visual concepts. This highlights the need for our canvas design to sample a few informative visual concepts. Next, we visualize the extraction of these key concepts through edge generation on this canvas. Finally, we demonstrate that \method captures surrounding concepts more accurately than BARON, which samples the surrounding area with a fixed window size.

\noindent \textbf{Canvas visualization.}
We consider the RPN results as candidate visual concepts, given their ability to detect novel categories in the image. As shown in \cref{fig:sup_main}-a, the top-$k$ red RPN boxes illustrate regions of potential objects within the image. Here, the $k$ is set to 300 by default. These RPN boxes are somewhat noisy and overly abundant.
To address this, the traditional RPN applies a post-processing technique, Non-Maximum Suppression (NMS), to generate region proposals, represented as green boxes. However, since RPN generates results biased toward base categories, the region proposals predominantly represent these categories. To address this, we design a probabilistic search algorithm to extract a few informative RPN boxes, including those for novel categories within the image.

To achieve this, we introduce a new data structure called the canvas. This structure converts the given image into a mesh grid, where each coordinate represents the probability of finding RPN boxes in the up, down, left, or right cardinal directions. Initially, each coordinate is assigned an equal probability in all directions, represented in dark purple in \cref{fig:sup_main}-b. To accelerate canvas construction, we reduce the top-$k$ RPN boxes to a small number of non-overlapping RPN boxes, as described in \cref{sec:sup:s1:overlap}-b. These non-overlapping RPN boxes are shown as light purple boxes. During probabilistic exploration across the coordinates, encountering these light purple areas redirects movement toward directions with key concepts.

\noindent \textbf{Edge visualization.}
We aim to extract key RPN boxes from the numerous RPN boxes. We utilize the key RPN boxes as visual concepts in our pipeline. To achieve this, we use the previously introduced canvas, which stores the probability of RPN box presence in neighboring directions at each coordinate. We demonstrate that \method effectively samples a few informative RPN boxes around each region proposal on this canvas.

Specifically, we create a subset that includes the region proposals extracted by RPN after NMS, along with $N$ extra region proposals. These extra region proposals are sampled from top-$k$ RPN boxes, chosen as those farthest from all other region proposals on average.
More details are in \cref{sec:sup:s1:fartest}. In \cref{fig:sup_main}-c, region proposals in this subset are represented by green boxes, while extra region proposals are shown as yellow boxes. Then, we generate edges connecting different pairs of region proposals in this subset. The edges detect the locations of key RPN boxes between pairs. As shown in \cref{fig:sup_main}-c, edges are represented by light blue boxes, with an interval $\Delta$ of 10 for clear visualization. Through probabilistic exploration between all pairs, we find that informative RPN boxes accumulate between these pairs. This results in converging on a few RPN boxes representing multiple meaningful concepts within the image. For example, in the ``traffic\_light'' case in \cref{fig:sup_main}-c, \method avoids sampling redundant RPN boxes in the sky area; instead, edges converge toward the traffic light area, where significant RPN boxes are densely clustered. In our pipeline, we leverage these key RPN boxes as visual concepts.

\noindent \textbf{Visualization comparison: \method vs. BARON.}
We visualize the region crops extracted from each bag of BARON and \method. Compared to BARON, we show that \method more accurately represents surrounding concepts, including novel categories. This demonstrates that \method achieves improved performance over BARON by leveraging an enhanced compositional structure.

Specifically, we observe that BARON's bag does not effectively capture surrounding novel objects due to its fixed window size for sampling neighboring regions, not concepts. In contrast, \method accurately represents surrounding novel categories by sampling neighboring areas based on the previously derived knowledge of key concepts. As shown in \cref{fig:sup_main}-d, the black dotted box represents the region crop within BARON's bag, while the blue box represents the view within \method's bag. For example, in the ``dining\_table'' case, \method captures the novel category of a dining table near the region proposal for a chair, which BARON often misses. Notably, \method works effectively in cases where the RPN frequently predicts small bounding boxes as region proposals. For example, in the ``airplane'' example, where the region proposal is located within the novel category area of an airplane, \method effectively represents the airplane area through the bag of views. 
Furthermore, we observe that \method also achieves an improved aspect ratio correction compared to BARON, as it samples based on a weighted sum of distance and aspect ratio.



\subsection{Detection Results}
\label{sec:sup:s4}
\begin{figure*}[t]
  \centering
   \includegraphics[width=\textwidth, height=0.95\textheight, keepaspectratio]{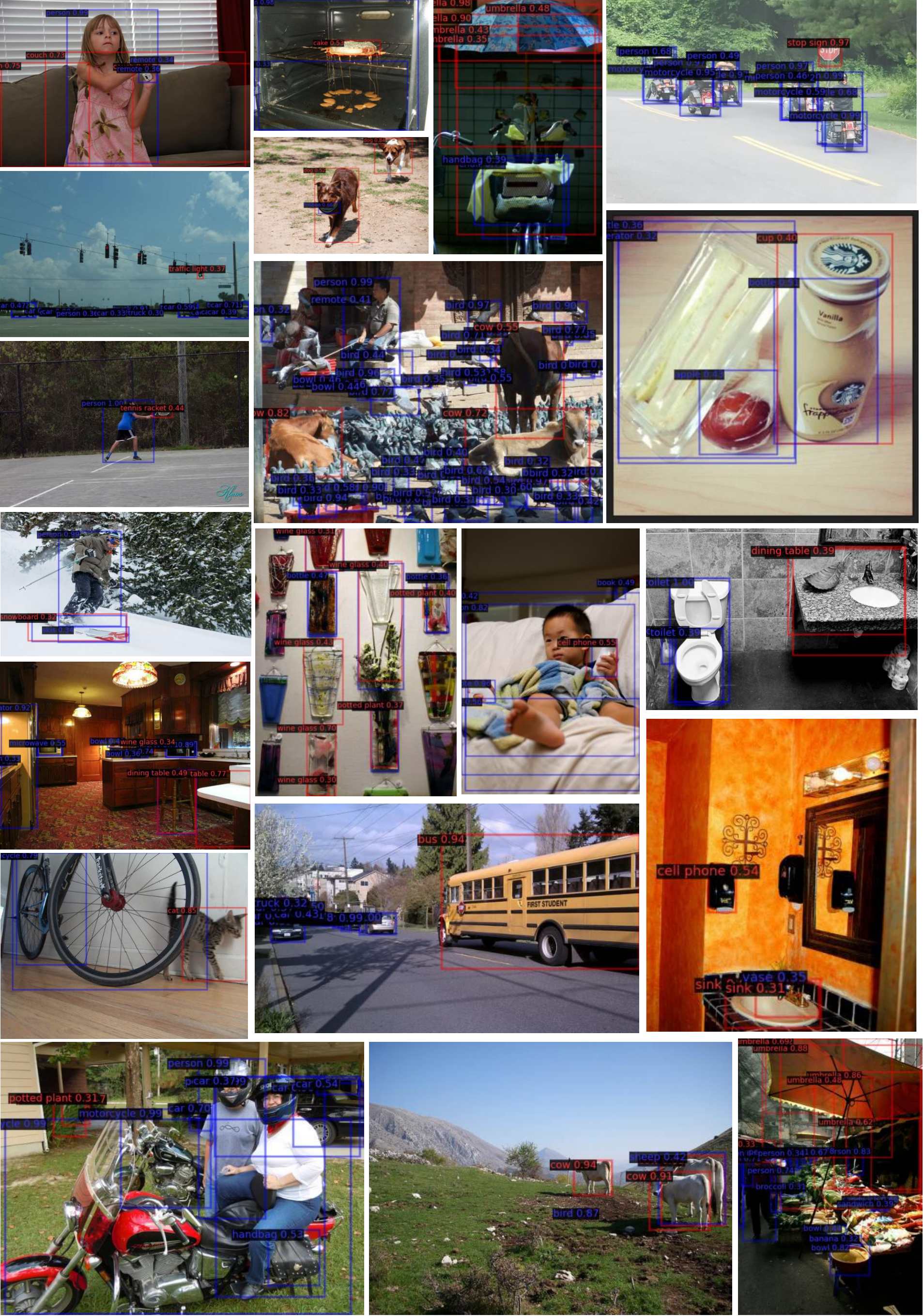}
   \caption{\textbf{Visualization of detection results on the OV-COCO dataset}. Red boxes and masks represent novel categories, while blue boxes and masks represent base categories.}
   \label{fig:sup_det_coco}
\end{figure*}

\begin{figure*}[t]
  \centering
   \includegraphics[width=\textwidth, height=0.95\textheight, keepaspectratio]{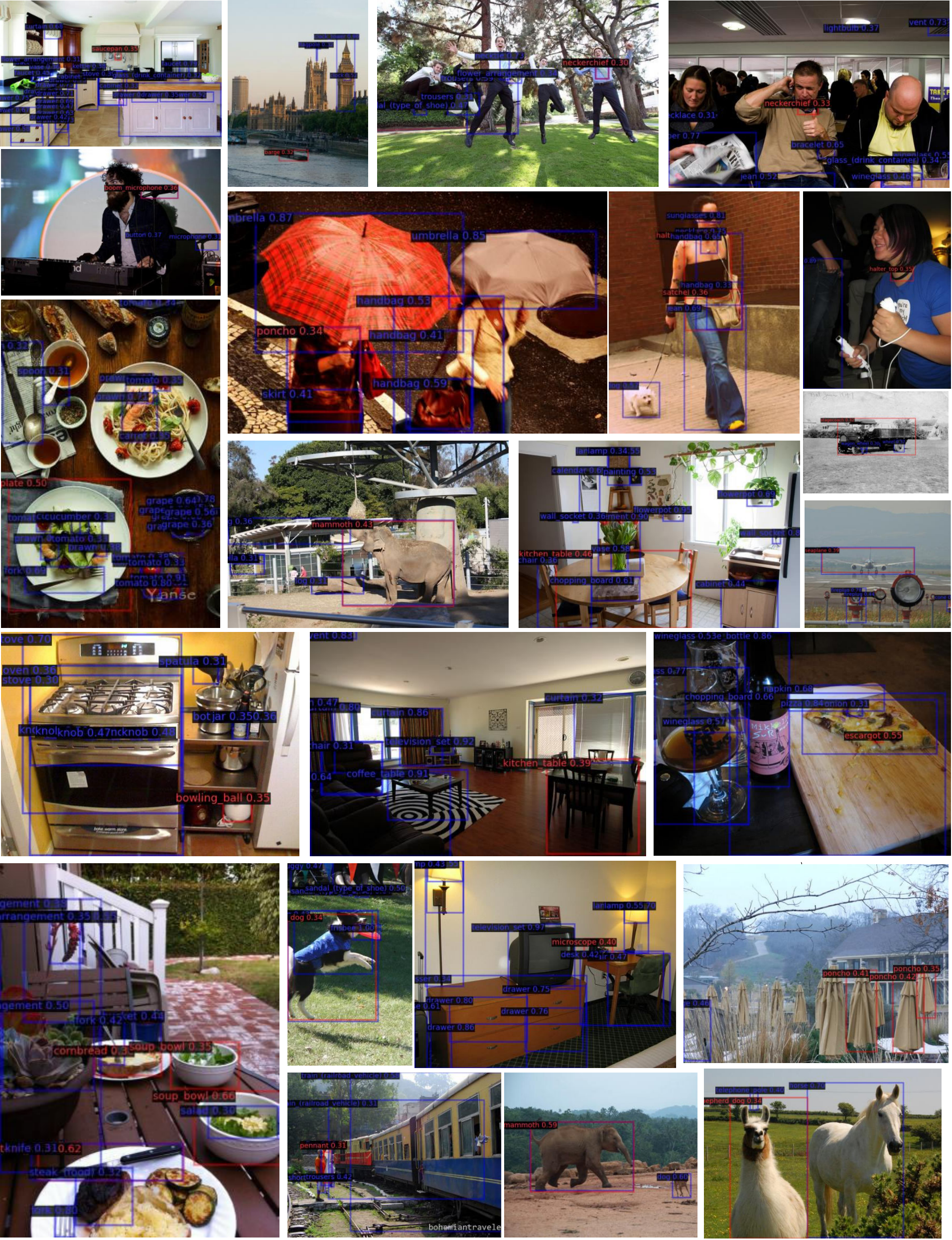}
   \caption{\textbf{Visualization of detection results on the OV-LVIS dataset}. Red boxes and masks represent novel (rare) categories, while blue boxes and masks represent base categories.}
   \label{fig:sup_det_lvis}
\end{figure*}

We show additional detection results from our method \method on two OVD benchmarks, OV-COCO and OV-LVIS, as shown in \cref{fig:sup_det_coco} and \cref{fig:sup_det_lvis}. The images are sourced from the benchmark validation set. On the COCO dataset, \method accurately detects novel categories, such as traffic light, bus, keyboard, cup, snowboard, and cow. For the LVIS dataset, \method identifies rare categories like boom\_microphone, mammoth, kitchen\_table poncho, escargot, shepherd\_dog, and pennant. Our findings show that \method accurately recognizes a wide range of novel objects defined in both benchmarks.


\end{document}